\documentclass[runningheads]{llncs}
\usepackage{graphicx}

\usepackage{tikz}
\usepackage{comment}
\usepackage{amsmath,amssymb} 
\usepackage{color}

\usepackage{multirow}
\usepackage{array}
\usepackage{booktabs}

\usepackage{animate}
\usepackage{lipsum}
\usepackage[misc]{ifsym}
\usepackage{hyperref}
\hypersetup{hidelinks,
	colorlinks=true,
	allcolors=green,
	pdfstartview=Fit,
	breaklinks=true}

\usepackage[accsupp]{axessibility}  

\begin{document}
\pagestyle{headings}
\mainmatter
\def\ECCVSubNumber{3887}  

\title{CCPL: Contrastive Coherence Preserving Loss for Versatile Style Transfer} 


\titlerunning{CCPL: Contrastive Coherence Preserving Loss for Versatile Style Transfer}

\author{Zijie Wu\inst{1}$^{*}$ \and
Zhen Zhu\inst{2}$^{*}$ \and
Junping Du\inst{3} \and
Xiang Bai\inst{1}$^{\dagger}$}

\authorrunning{Wu et al.}

\institute{Huazhong University of Science and Technology, China \and
University of Illinois at Urbana-Champaign, USA \and
Beijing University of Posts and Telecommunications, China \\
\email{\{zijiewu,xbai\}@hust.edu.cn,
zhenzhu4@illinois.edu,
junpingdu@126.com}}
\maketitle

\newcommand\blfootnote[1]{%
\begingroup
\renewcommand\thefootnote{}\footnote{#1}%
\addtocounter{footnote}{-1}%
\endgroup
}

\begin{abstract}
In this paper, we aim to devise a universally versatile style transfer method capable of performing artistic, photo-realistic, and video style transfer jointly, without seeing videos during training. Previous single-frame methods assume a strong constraint on the whole image to maintain temporal consistency, which could be violated in many cases. Instead, we make a mild and reasonable assumption that global inconsistency is dominated by local inconsistencies and devise a generic Contrastive Coherence Preserving Loss (CCPL) applied to local patches. CCPL can preserve the coherence of the content source during style transfer without degrading stylization. Moreover, it owns a neighbor-regulating mechanism, resulting in a vast reduction of local distortions and considerable visual quality improvement. Aside from its superior performance on versatile style transfer, it can be easily extended to other tasks, such as image-to-image translation. Besides, to better fuse content and style features, we propose Simple Covariance Transformation (SCT) to effectively align second-order statistics of the content feature with the style feature. Experiments demonstrate the effectiveness of the resulting model for versatile style transfer, when armed with CCPL. 

\keywords{image style transfer, video style transfer, temporal consistency, contrastive learning, image-to-image translation.}
\end{abstract}

\blfootnote{$*$ Equal contribution ~~~~~ $\dagger$ Corresponding author}

\begin{figure}[t]
\centering
\animategraphics[autoplay,loop,width=1.0\textwidth]{6}{Fig1/pic}{1}{10}
\caption{Our algorithm can perform versatile style transfer. From left to right are examples of artistic  image/video style transfer, photo-realistic image/video style transfer. Adobe Acrobat Reader is recommended to see the animations.}
\label{fig1}
\end{figure}

 \section{Introduction}
\label{intro}

Over the past years, much progress has been made on style transfer to make the result exceptionally pleasant and artistically valuable. In this work, we are interested in \emph{versatile style transfer}. Apart from artistic style transfer and photo-realistic style transfer, our derived method is versatile in performing video style transfer well without explicitly training with videos. The code is available at \href{https://github.com/JarrentWu1031/CCPL}{\textcolor{red}{https://github.com/JarrentWu1031/CCPL}}.

One naive solution to produce a stylized video is to independently transfer the style of successive frames with the same style reference. Since no temporal consistency constraint is enforced, the generated video usually has obvious flicker artifacts and incoherence between two consecutive frames. To combat this problem, former methods~\cite{chen2017coherent,gao2018reconet,gupta2017characterizing,huang2017real,ruder2016artistic,ruder2018artistic} used optical flow as guidance to restore the estimated motions of the original videos. However, estimating optical flow requires much computation, and the accuracy of estimated motions tightly constrains the quality of the stylized video. Recently, some algorithms~\cite{deng2020arbitrary,li2019learning,liu2021adaattn} tried to improve the temporal consistency of the video outputs with single-frame regularizations. They attempted to ensure a linear transformation from the content feature to the fused feature. The underlying idea is to encourage preserving the dense pairwise relations within the content source. However, without explicit guidance, the linearity is largely affected by the global style optimization. Therefore, their video results are still not that temporally consistent. We notice that most video results show good structure rigidity to their content video inputs, but the local noise escalates the impression of inconsistency. So instead of considering a global constraint that could be easily violated, we start by thinking about a more relaxed constraint defined on local patches.

\begin{figure}[t]
\centering
\includegraphics[width=1.0\textwidth]{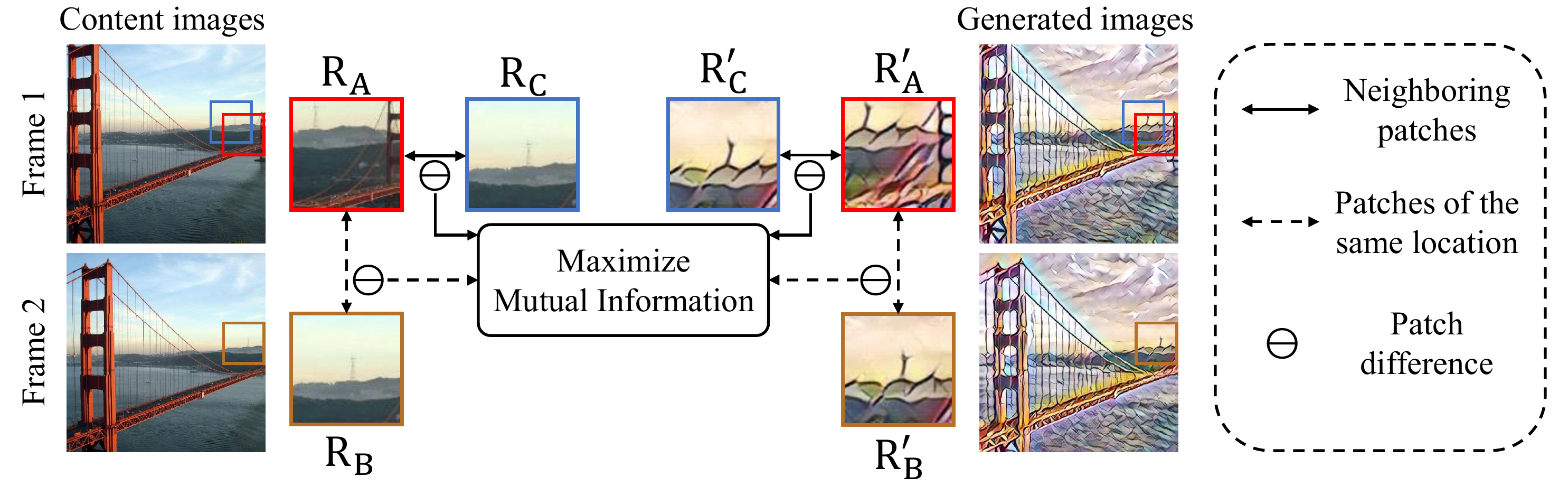} 
\caption{Intuition of the contrastive coherence preserving loss. The regions denoted with \textcolor{red}{red} boxes from the first frame ($\mathrm{R_{A}}$ or $\mathrm{R_{A}^{'}}$) have the same location with corresponding patches in the second frame wrapped with \textcolor{brown}{brown} box ($\mathrm{R_{B}}$ or $\mathrm{R_{B}^{'}}$). $\mathrm{R_{C}}$ and $\mathrm{R_{C}^{'}}$ (in the \textcolor{blue}{blue} boxes) are cropped from the first frames but their semantics align with $\mathrm{R_{B}}$ and $\mathrm{R_{B}^{'}}$. The difference between two patches is denoted as $\mathcal{D}$ ($e.g.$, $\mathcal{D}\mathrm{(R_{A},R_{B})}$). The mutual information between $\mathcal{D}\mathrm{(R_{A},R_{C})}$ and $\mathcal{D}\mathrm{(R_{A}^{'},R_{C}^{'})}$ ($\mathcal{D}\mathrm{(R_{A},R_{B})}$ and $\mathcal{D}\mathrm{(R_{A}^{'},R_{B}^{'})}$) is encouraged to be maximized to preserve the coherence of the content source.}
\label{fig2}
\end{figure}

As shown in Fig.~\ref{fig2}, our idea is simple: the change between patches denoted by $\mathrm{R_{A}^{'}}$ and $\mathrm{R_{B}^{'}}$ of the same location in the stylized images should be similar to patches $\mathrm{R_{A}}$ and $\mathrm{R_{B}}$ of two adjacent content frames. If the two consecutive content frames are shot within a short period, it is likely to find a similar patch to $\mathrm{R_{B}}$ in the neighboring area, which is denoted by $\mathrm{R_{C}}$ (in the blue box). In other words, 
we can treat two nearby patches in the same image as patches of the same location in consecutive frames. Therefore, we can apply the constraint even when we only have single-frame images. However, forcing these patch differences to be the same is unreliable since it will encourage the outputs to be the same as the content images. Then no style transfer effects would appear in the results. Inspired by recent advances in contrastive learning~\cite{chen2020simple,oord2018representation,park2020contrastive}, we use the InfoNCE loss~\cite{oord2018representation} to maximize the mutual information between the positive pair (from the same region) of patch differences relative to other negative pairs (from different regions). By sampling a sufficient number of negative pairs, the loss encourages the positive pair to be close while keeping away from negative samples. We call the derived loss as~\textbf{Contrastive Coherence Preserving Loss (CCPL)}.

After applying CCPL, we note that the temporal consistency of the video outputs improves substantially while the stylization remains satisfying (see Fig.~\ref{fig5} and Tab.~\ref{table1}). Besides, due to the neighbor-regulating strategy of the CCPL, the local patches of the generated image are constrained by their neighboring patches, which reduces local distortions significantly, thus leading to better visual quality. Our proposed CCPL does not require video inputs and is not bound to specific network architecture. Therefore we can apply it to any existing image style transfer networks during training to improve their performance on images and videos (see Fig.~\ref{fig9} and Tab.~\ref{table1}). The significant improvement in visual quality and its flexibility empowers CCPL for photo-realistic style transfer with minor modifications, marking it a vital tool towards versatile style transfer (see Fig.~\ref{fig1}).

With CCPL, we now aspire to fuse content and style features both efficiently and effectively. To realize this, we propose an efficient network for versatile style transfer, called \textbf{SCTNet}. The critical element of SCTNet is the \textbf{Simple Covariance Transformation (SCT)} module to fuse style features and content features. It computes the covariance of the style feature and directly multiplies the feature covariance with the normalized content features. Compared to the fusing operations in AdaIN~\cite{huang2017arbitrary} and Linear~\cite{li2019learning}, our SCT is simple and can capture precise style information at the same time. 

To summarize, our contributions are three-fold:

\begin{enumerate}
    \item We propose Contrastive Coherence Preserving Loss (CCPL) for versatile style transfer. It encourages consistency between the content image and generated image in terms of the difference of an image patch with its neighboring patches. It is effective and transferable to other style transfer methods.
    \item We propose Simple Covariance Transformation (SCT) to align second-order statistics of content and style features effectively. The resulted SCTNet is structurally simple and remains efficient (about 25 frames per second at the scale of $512\times512$), which is of great potential for practical use.
    \item We apply our CCPL to other tasks, such as image-to-image translation, and improve the temporal consistency and visual quality of results without further modifications, demonstrating the flexibility of CCPL.
\end{enumerate}

\section{Related Works}
\textbf{Image Style Transfer.}~These algorithms aim at generating an image with the structure of one image and the style of another. Gatys $et~al.$ first pioneered Neural Style Transfer (NST)~\cite{gatys2016image}. For acceleration, some algorithms~\cite{johnson2016perceptual,ulyanov2016texture} approximated the iterative optimization procedure as feed-forward networks and achieved style transfer with a fast forward pass. For broader applications, several algorithms tried to transfer multiple styles within a single model~\cite{chen2017stylebank,dumoulin2016learned}. Nevertheless, these models have limitations on the number of learnt styles. Since then, various methods have been designed to transfer style from random images.

Style-swap methods~\cite{chen2016fast,sheng2018avatar} swapped each content patch with its closest style patch before reconstructing the image. WCT~\cite{li2017universal} utilized singular value decomposition to whiten and then re-color images. AdaIN~\cite{huang2017arbitrary} replaced the feature means and standard deviations with those from the style source. Recently, many attention-based algorithms came forth. For example, Li $et~al.$~\cite{li2019learning} devised a linear transformation to align second-order statistics between the fused feature and the style feature. Deng $et~al.$~\cite{deng2020arbitrary} improved it with multi-channel correlating. SANet~\cite{park2019arbitrary} re-arranged style features utilizing spatial correlations with content features. AdaAttN~\cite{liu2021adaattn} combined AdaIN~\cite{huang2017arbitrary} and SANet~\cite{park2019arbitrary} to balance global and local style effects. Cheng $et~al.$~\cite{cheng2021style} proposed style-aware normalized loss to balance stylization. Another branch aims to transfer photo-realistic style onto images. Luan $et~al.$~\cite{luan2017deep} designed a color transformation network inspired by the Matting Laplacian~\cite{levin2007closed}. Li $et~al.$~\cite{li2018closed} replaced the upsampling layers of WCT~\cite{li2017universal} with unpooling layers and added max-pooling masks to alleviate detail losses. Yoo $et~al.$~\cite{yoo2019photorealistic} introduced the wavelet transform to preserve structural information. An $et~al.$~\cite{an2019stylenas} used neural architecture search algorithms to find the appropriate decoder design for better performance.

\vspace{2mm} \noindent \textbf{Video Style Transfer.}~Existing video style transfer algorithms can be roughly divided into two categories according to whether to use the optical flow or not.

One line of work leverages optical flow when producing the video output. These algorithms try to estimate the motion of the original video and restore it in the generated video. Ruder $et~al.$~\cite{ruder2016artistic} proposed a temporal loss to regulate the current frame with the warped previous frame to extend the image style transfer algorithm~\cite{gatys2016image} to videos. Chen $et~al.$~\cite{chen2017coherent} designed an RNN structure baseline and performed the warping operation in the feature domain. Gupta $et~al.$~\cite{gupta2017characterizing} concatenated the former stylized frame with the current content frame before rendering and formed a flow loss as a constraint. Huang $et~al.$~\cite{huang2017real} tried to integrate temporal coherence into the stylization network with a hybrid loss. Ruder $et~al.$~\cite{ruder2018artistic} extended their previous work~\cite{ruder2016artistic} with new initializations and loss functions to improve robustness against large motions and strong occlusions.  Temporal consistency can be improved with these optical flow constraints. However, optical flow estimation is not perfectly accurate, resulting in artifacts in the video results. Besides, it is computationally expensive, especially when the image size scales up. Considering these, another line of work tries to maintain the coherence of content inputs without using optical flow.

Li $et~al.$~\cite{li2019learning} and Deng $et~al.$~\cite{deng2020arbitrary} devised linear transformations for content features to preserve structure affinity. Liu $et~al.$~\cite{liu2021adaattn} used L1 normalization to replace the softmax operation of SANet~\cite{park2019arbitrary} to get a more flat attention score distribution. Wang $et~al.$~\cite{wang2020consistent} proposed compound temporal regularization to enhance the robustness of the network to motions and illumination changes. Compared to these approaches, our proposed CCPL poses no requirements for the network architecture, making it exceptionally adaptive to other networks. With our SCTNet, the temporal consistency of video outputs surpasses SOTAs while the stylization remains satisfying. We also apply CCPL to other networks. The results show similar improvements in video stability (see Tab.~\ref{table1}). 

\begin{figure}[t]
\centering
\includegraphics[width=1.0\textwidth]{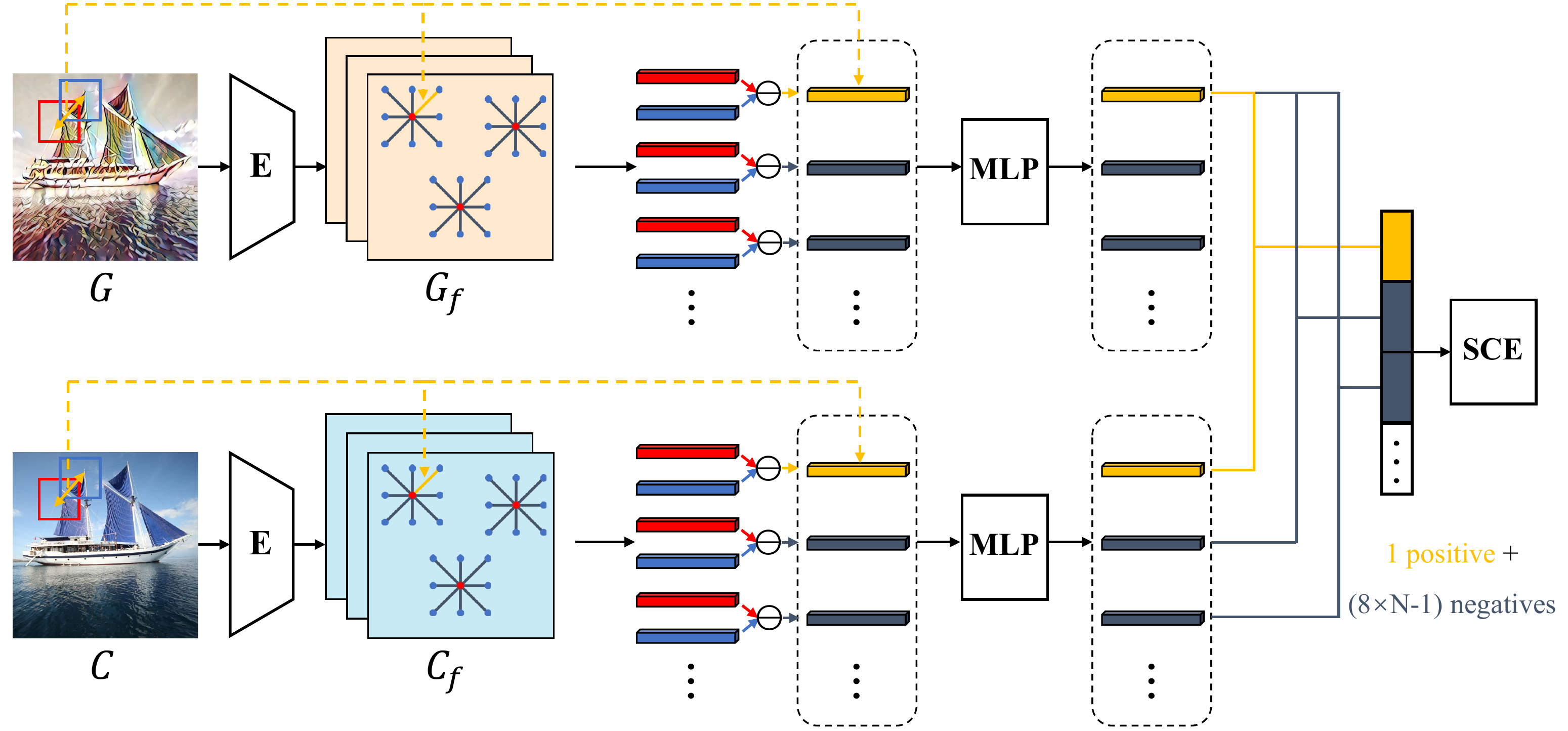} 
\caption{Diagram of the proposed CCPL. $C_{f}$ and $G_{f}$ represent the encoded features from a specific layer of the encoder $E$. $\ominus$ denotes vector subtraction, and SCE means softmax cross-entropy. The yellow dashed lines illustrate how the positive pair is produced.}
\label{fig3}
\end{figure}

\vspace{2mm} \noindent \textbf{Contrastive Learning.}~The original purpose of contrastive learning algorithms is to learn a good feature representation in a self-supervised scenario. A rich family of methods tried to achieve this by maximizing the mutual information of positive feature pairs while minimizing it in negative pairs~\cite{chen2020simple,chen2020improved,chen2021exploring,grill2020bootstrap,he2020momentum,oord2018representation}. Recent works extended contrastive learning to the field of image-to-image translation~\cite{park2020contrastive} and image style transfer~\cite{chen2021artistic}. Our work is most relevant to CUT~\cite{park2020contrastive} in using patch-based InfoNCE loss~\cite{oord2018representation}. But CUT~\cite{park2020contrastive} utilized the correspondence of patches at the same locations for the image-to-image (Im2Im) translation task. However, our CCPL incorporates a neighbor-regulating scheme to preserve the correlations among neighboring patches, making it suitable for image and video generation. Besides, our experiment illustrates the effectiveness of CCPL on top of CUT~\cite{park2020contrastive} in the Im2Im translation task, as depicted in Sec.~\ref{app}. 

\section{Methods}


\subsection{Contrastive Coherence Preserving Loss}
\label{ccp}

Given two frames $C_t$ and $C_{t+\Delta t}$ where $\Delta t$ is the time interval in between, we assume the difference between the corresponding generated images $G_t$ and $G_{t+\Delta t}$ is linearly dependent on the difference between $C_t$ and $C_{t+\Delta t}$, when $\Delta t$ is small:
\begin{equation}
\label{assumption}
    \lim_{\Delta t \rightarrow 0} \mathcal{D}(C_{t+\Delta t}, C_t) \simeq \mathcal{D}(G_{t+\Delta t}, G_t),
\end{equation}
where $\mathcal{D}(a, b)$ represents the difference between $a$ and $b$. 
This constraint is probably too strict to hold for the whole image but technically sound for local patches where usually only simple image transformations, \emph{e.g.}, translation or rotation, can occur. Under this assumption, we propose a generic Contrastive Coherence Preserving Loss (CCPL) applied to local patches to enforce this constraint. We show in Sec.~\ref{intro} that our loss applied on neighboring patches is equivalent to that on corresponding patches of two frames, assuming $\Delta t$ is small. Operating on a single frame frees us from processing multiple frames of a video source, saving computation budget. 

To apply CCPL, first, we send the generated image $G$ and its content input $C$ to the fixed image encoder $E$ to get feature maps of a specific layer, denoted as $G_f$ and $C_f$ (shown in Fig.~\ref{fig3}). Second, we randomly sample $N$ vectors\footnote{As encoded features are spatially decreased, each vector in the feature level corresponds to an image patch in the image level.} from $G_f$ ({red} dots in Fig.~\ref{fig3}), denoted as $G_{a}^{x}$ where $x=1, \cdots, N$. Third, we sample the \emph{eight} nearest neighboring vectors of each $G_{a}^{x}$ ({blue} dots in Fig.~\ref{fig3}), denoted by $G_{n}^{x,y}$ where $y=1, \cdots, 8$ is the neighbor index. Then, we accordingly sample from $C_f$ at the same locations to get $C_{a}^{x}$ and $C_{n}^{x,y}$, respectively. The differences between a vector and its neighboring vectors are measured by:
\begin{equation}
\label{equ1}
\begin{aligned}
    d_{g}^{x,y} = G_{a}^{x} \ominus G_{n}^{x,y}&,~ 
    d_{c}^{x,y} = C_{a}^{x} \ominus C_{n}^{x,y},
\end{aligned}    
\end{equation}
where $\ominus$ represents vector subtraction. In order to realize Eq.~\ref{assumption}, one simple thought is to enforce $d_{g}$ equal to $d_{c}$. But in this case, an easy workaround of the network is to encourage $G$ similar to $C$, meaning that this constraint would contradict the purpose of style transfer. Inspired by the recent progress in contrastive learning~\cite{chen2020simple,he2020momentum,oord2018representation}, we instead try to maximize the mutual information between ``positive" difference vector pairs. A pair is only defined between a difference vector from $C_f$ and $G_f$. Namely, the difference vectors of the same locations are defined as positive pairs between $d_{g}$ and $d_{c}$, otherwise negative. The underlying intuition is also straightforward: the difference vectors of the same location should be most relevant in the latent space compared to other random pairs.
 
We follow the design of~\cite{chen2020simple} to build a two-layer MLP (multi-layer perceptron) to map the difference vectors and normalize them onto a unit sphere before computing InfoNCE loss~\cite{oord2018representation}. Mathematically:
\begin{equation}
\label{equ2}
L_{\mathrm{ccp}} = \sum_{m=1}^{8\times N}-\log[\frac{\exp(d_{g}^{m} \cdot d_{c}^{m} /\tau )}{\exp(d_{g}^{m} \cdot d_{c}^{m}/\tau ) + \sum_{n=1,n\ne~m}^{8\times N}\exp(d_{g}^{m} \cdot d_{c}^{n}/\tau )}],   
\end{equation}
where $\tau$ stands for a temperature hyper-parameter set to 0.07 by default.
With this setting, the temporal consistency of video outputs improves significantly (see Fig.~\ref{fig5} and Tab.~\ref{table1}) while the stylization remains satisfying or even gets better (see Fig.~\ref{fig6}, Fig.~\ref{fig9}, dirty texture disappears with our CCPL).

This loss avoids direct contradiction with style losses used to ensure style coherence between the generated and style image. Meanwhile, it can improve the temporal consistency of the generated video even without leveraging information from other frames of the input video. The complexity of CCPL is $\mathcal{O}\left ( 8\times N  \right )^{2}$, where $8\times N$ represents the number of sampled difference vectors. It is computationally affordable during training and has zero influence on inference speed (shown in Fig.~\ref{fig8}$a$). CCPL can even work as a simple plugin to extend methods of other \emph{image generation} tasks to produce videos with much better temporal consistency, as shown in Sec.~\ref{app}. 

\begin{figure}[t]
\centering
\includegraphics[width=1.0\textwidth]{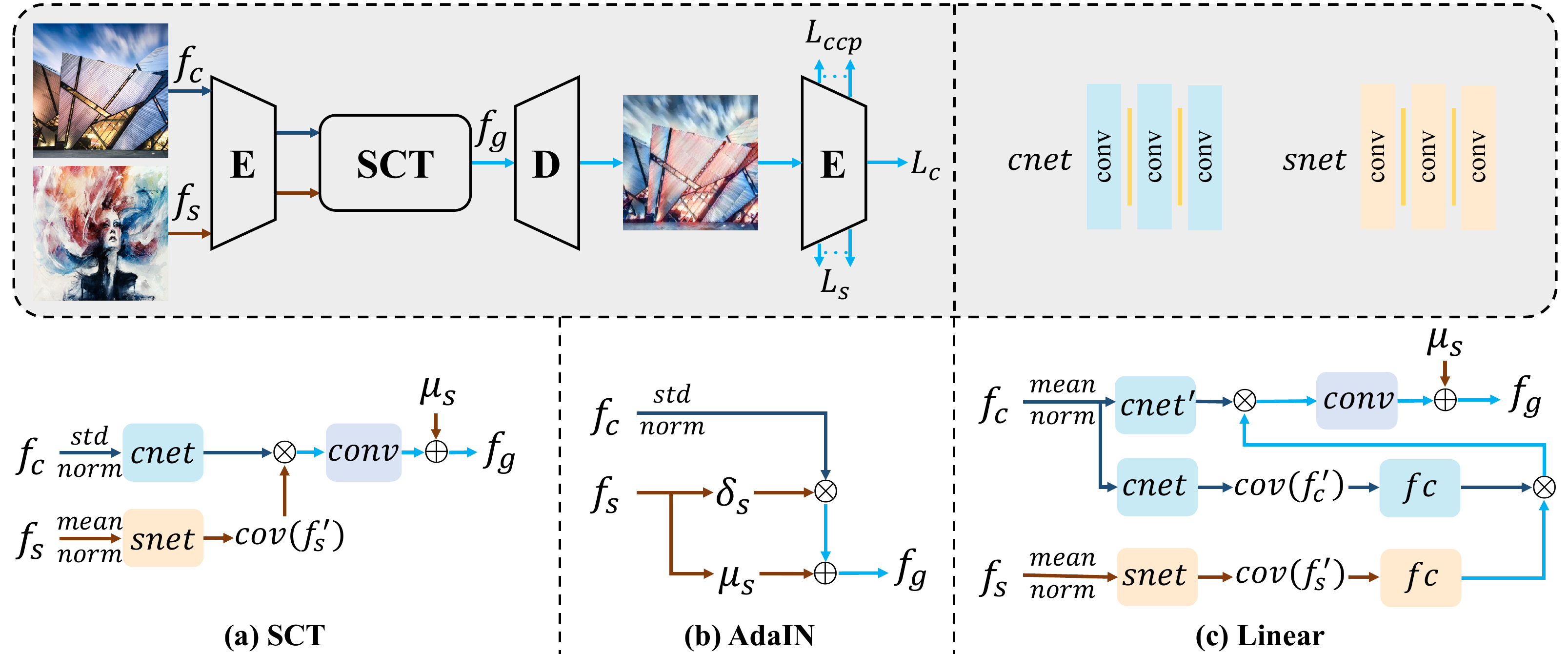} 
\caption{Details of the proposed SCT module and its comparison with similar algorithms (AdaIN~\cite{huang2017arbitrary}, Linear~\cite{li2019learning}). Here $conv$ represents a convolutional layer, and the yellow lines in $cnet$ and $snet$ denote $relu$ layers. Besides, $std~norm$ represents normalizing features by the means and standard deviations of channels, while $mean~norm$ normalizes features by the means of its channels.}
\label{fig4}
\end{figure}


\subsection{Simple Covariance Transformation}
\label{SCT}
With CCPL guaranteeing temporal consistency, our next goal is to design a simple and effective module for the fusion of content and style features for rich stylization. Huang $et~al.$~\cite{huang2017arbitrary} proposed AdaIN to align channel-wise mean and variance of content and style features directly. Although simple enough, the inter-channel correlations are ignored, which are verified to be effective in the latter literature~\cite{deng2020arbitrary,li2019learning}. Li $et~al.$~\cite{li2019learning} devised a channel-attention mechanism to transfer second-order statistics of style features onto corresponding content features. But we empirically find that the structure of Linear~\cite{li2019learning} can be simplified.

To combine the advantages of AdaIN~\cite{huang2017arbitrary} and Linear~\cite{li2019learning}, we design a \textbf{Simple Covariance Transformation (SCT)} module to fuse style and content features. As shown in Fig.~\ref{fig4}, first, we normalize the content feature $f_{c}$ by the means and deviations of its channels~\cite{huang2017arbitrary} and the style feature $f_{s}$ by the means of its channels~\cite{li2019learning} to get $\bar{f_{c}}$ and $\bar{f_{s}}$. To reduce computation costs, we send $\bar{f_{c}}$ and $\bar{f_{s}}$ to $cnet$ and $snet$ ($cnet$ and $snet$ both contain three convolutional layers, and two $relu$ layers in between) to gradually reduce the dimension of channels $(512\rightarrow32)$, and get $f_{c}^{'}$ and $f_{s}^{'}$. Then we flatten $f_{s}^{'}$ and calculate its covariance matrix $cov(f_{s}^{'})$ to find out the channel-wise correlations. After that, we simply fuse the features by performing a matrix multiplication between $cov(f_{s}^{'})$ and $f_{c}^{'}$ to get $f_{g}$. Finally, we use a single convolutional layer (denoted as $conv$ in Fig.~\ref{fig4}) to restore the channel dimension of $f_{g}$ back to normal ($32\rightarrow512$) and add channel means of the original style feature before sending it to the decoder. 

Combined with a symmetric encoder-decoder module, we name the whole network as \textbf{SCTNet}. The encoder is a VGG-19 network~\cite{simonyan2014very} pre-trained on ImageNet~\cite{deng2009imagenet} to extract features from the content and style images, while the symmetric decoder needs to convert the fused feature back to images. Experiments suggest that our SCTNet is comparable to Linear~\cite{li2019learning} in stylization effects (see Fig.~\ref{fig6} and Tab.~\ref{table1}), while being lighter and faster (see Tab.~\ref{table3}).

\subsection{Loss Function}
Apart from the proposed CCPL, we adopt two commonly used losses~\cite{an2021artflow,deng2020arbitrary,huang2017arbitrary,liu2021adaattn} for style transfer. The overall training loss is a weighted sum of these three losses:
\begin{equation}
\label{equ7}
L_{\mathrm{totoal}} = \lambda _{c}\cdot L_{\mathrm{c}} + \lambda _{s}\cdot L_{\mathrm{s}} + \lambda _{ccp}\cdot L_{\mathrm{ccp}}.
\end{equation}

The content loss $L_c$ (the style loss $L_{s}$) is measured by the Frobenius norm of the differences between (means $\mu(\cdot)$ and standard deviations $\sigma(\cdot)$ of) the generated features and the content (style) features:  
\begin{equation}
\label{equ8}
\begin{aligned}
L_{\mathrm{c}} &= \left \| \phi_{l}(I_{g}) -  \phi_{l}(I_{c})\right \|_{F},
\end{aligned}
\end{equation}
\begin{equation}
\label{equ9}
\begin{aligned}
L_{\mathrm{s}} = \sum_{l}(\left \| \mu(\phi_{l}(I_{g})) - \mu (\phi_{l}(I_{s}))\right \|_{F}  +  
\left \| \sigma(\phi_{l}(I_{g})) - \sigma(\phi_{l}(I_{s})) \right \|_{F}),
\end{aligned}
\end{equation}
where $\phi_{l}(\cdot)$ denotes the feature map from the $l$-th layer of the encoder. For artistic style transfer, we use the features from \{$relu4\_1$\}, \{$relu1\_1$, $relu2\_1$, $relu3\_1$, $relu4\_1$\}, \{$relu2\_1$, $relu3\_1$, $relu4\_1$\} to calculate the content loss, style loss, and CCPL, respectively. As for photo-realistic style transfer, we set the loss layers to \{$relu3\_1$\}, \{$relu1\_1$, $relu2\_1$, $relu3\_1$\}, \{$relu1\_1$, $relu2\_1$, $relu3\_1$\} for the above losses. The loss weights are set to $\lambda_{c}=1.0,~\lambda_{s}=10.0,~\lambda_{ccp}=5.0$ by default. Please check Sec.~\ref{ablations} for details about how we find these configurations.

\section{Experiments}

\subsection{Experimental settings}
\label{imp}
\textbf{Implementation details.} We adopt content images from MS-COCO~\cite{lin2014microsoft} data-set and style images from Wikiart~\cite{phillips2011wiki} data-set to train our network. Both data-sets contain approximately 80,000 images. We use the Adam optimizer~\cite{kingma2014adam} with a learning rate of 1e-4 and the batch size of 8 to train the model for 160k iterations by default. During training, we first resize the smaller dimension of images to 512. Then we randomly crop $256\times256$ patches from images as the final input. For CCPL, we only treat difference vectors within the same content image as negative samples. More details are provided in the supplemental file. 

\begin{table}[t]
\centering
\caption{Quantitative comparison of video and artistic style transfer. Here $i$ stands for the interval of frames, and $\mathrm{Pre.}$ stands for \emph{human~preference~score}. We show the \emph{human~preference~score} of both artistic image style transfer (Art) and video style transfer (Vid) in the table. The results of \emph{temporal~loss} are magnified $100$ times. We show the \textbf{first-place} score in bold and the \underline{second-place} score with underlining.}
\begin{tabular}{l|p{1.5cm}<{\centering}|*{2}{p{1.2cm}<{\centering}}|*{2}{p{1.2cm}<{\centering}}|*{2}{p{1.2cm}<{\centering}}}
\toprule[1pt]
\multirow{2}{*}{Methods} & \multirow{2}{*}{SIFID ($\downarrow$)} & \multicolumn{2}{c|}{LPIPS($\downarrow$)} & \multicolumn{2}{c|}{Temporal Loss ($\downarrow$)} & \multicolumn{2}{c}{Pre. ($\uparrow$)}\\
\cline{3-4} \cline{5-6} \cline{7-8}
 &   & i=1  & i=10  & i=1 & i=10 & Art & Vid \\
    \hline 
    AdaIN~\cite{huang2017arbitrary} & 2.44 & 0.184 & 0.444 & 5.16 & 7.92 & 0.028 & 0.028\\
    AdaIN~\cite{huang2017arbitrary}+$L_{\mathrm{ccp}}$ & 2.58 & 0.163 & 0.408 & 4.21 & 6.72 & 0.054 & 0.054\\
    SANet~\cite{park2019arbitrary} & 2.40 & 0.227 & 0.478 & 6.31 &  13.72 & 0.062 & 0.046\\
    SANet~\cite{park2019arbitrary}+$L_{\mathrm{ccp}}$ & 2.60 & 0.167 & 0.390 & 4.42 & 7.09 & 0.084 & 0.086\\
    Linear~\cite{li2019learning} & 2.38 & 0.160 & 0.417 & 4.25 & 7.61 & 0.076 & 0.080\\
    Linear~\cite{li2019learning}+$L_{\mathrm{ccp}}$ & 2.47 & 0.147 & 0.370 & 4.01 & 6.96 & 0.082 & 0.088\\
    MCCNet~\cite{deng2020arbitrary} & 2.34 & 0.162 & 0.424 & 4.21& 7.64 & 0.088 & \underline{0.106}\\
    AdaAttN~\cite{liu2021adaattn} & 2.48 & 0.207 & 0.419 & 4.87& 6.49 & \underline{0.098} & 0.094\\
    DSTN~\cite{hong2021domain} & 2.83 & 0.234 & 0.450 & 5.72 & 10.76 & 0.070 & 0.038\\
    IE~\cite{chen2021artistic} & 2.99 & 0.182 & 0.379 & 4.35 & 6.76 & 0.054 & 0.058\\
    \hline
    ReReVST~\cite{wang2020consistent} & 2.78 & \textbf{0.137} & \textbf{0.359} & \textbf{2.97} & \underline{5.19} & 0.046 & 0.062\\
    \hline
    SCTNet & \textbf{2.29} & 0.187 & 0.446 & 4.82 & 12.22 & 0.066 & 0.060\\
    SCTNet+$L_{\mathrm{nor}}$\cite{cheng2021style} & \underline{2.31} & 0.191 & 0.439 & 5.07 & 11.54 & 0.070 & 0.062\\
    SCTNet+$L_{\mathrm{ccp}}$ & 2.43 & \underline{0.144} & \underline{0.367} & \underline{3.45} & \textbf{5.08} & \textbf{0.122} & \textbf{0.138}\\
\bottomrule[1pt] 
\end{tabular} 
\label{table1}
\end{table}

\begin{table}[t]
\centering
\caption{Quantitative comparison of photo-realistic style transfer.}
\begin{tabular}{l|c|c|c|c|c|c}
\toprule[1pt]
    Metrics & Linear~\cite{li2019learning} & $\mathrm{WCT^{2}}$ ~\cite{yoo2019photorealistic} & StyleNAS~\cite{an2019stylenas} & DSTN~\cite{hong2021domain} & SCTNet & SCTNet+$L_{\mathrm{ccp}}$\\
    \hline 
    SIFID ($\downarrow$) & 1.82 & 1.86 & 2.37 & 3.35 & \textbf{1.65} & 2.14  \\
    LPIPS ($\downarrow$) & 0.395 & 0.419 & 0.379 & 0.464 & 0.427 & \textbf{0.351}  \\
    Pre. ($\uparrow$) & 0.176 & 0.186 & 0.180 & 0.068 & 0.128 & \textbf{0.262}  \\
\bottomrule[1pt]
\end{tabular}
\label{table2}
\end{table}

\vspace{2mm} \noindent\textbf{Metrics.}
To comprehensively evaluate the performance of different algorithms and make the comparison fair, we adopt several metrics to assess the results' stylization effects and temporal consistency. To evaluate stylization effects, we compute \emph{SIFID}~\cite{shaham2019singan} between the generated image and its style input to measure their style distribution distance. Lower SIFID represents closer style distributions of a pair. To evaluate the visual quality and temporal consistency, we opt to \emph{LPIPS}~\cite{zhang2018unreasonable}, which is originally used to measure the diversity of the generated images~\cite{choi2020stargan,huang2018multimodal,lee2018diverse}. In our cases, small LPIPS represents few local distortions of the photo-realistic results or minor changes between two stylized video frames. Nonetheless, LPIPS only considers the correlations between stylized video frames while ignoring the changes between the original frames. As a supplement, we also adopt the \emph{temporal loss} defined in~\cite{wang2020consistent} to measure temporal consistency. It is done by utilizing the optical flow between two frames to warp one stylized result and compute the Frobenius difference with another. We evaluate short-term (two adjacent frames) and long-term (9 frames in between) consistency for video style transfer. For short-term consistency, we directly use the ground-truth optical flow from the MPI Sintel data-set~\cite{butler2012naturalistic}. Otherwise, we use PWC-Net~\cite{sun2018pwc} to estimate the optical flow between two frames. The lower temporal loss represents better preservation of coherence between two frames.

For image style transfer comparison, we randomly choose 10 content images and 10 style images to synthesize 100 stylized images for each method and calculate their mean SIFID as the stylization metric. Besides, we compute the mean LPIPS to measure the visual quality of photo-realistic results. As for temporal consistency, we randomly select 10 video clips (50 frames, 12 FPS each) from the MPI Sintel dataset~\cite{butler2012naturalistic} and use 10 style images to transfer these videos, respectively. Then we compute the mean LPIPS and temporal loss as the temporal consistency metrics. We also include human evaluation, which is more representative in image generation tasks. To do so, we invite 50 participants to choose their favorite stylized image/video from each image/video-style pair considering the visual quality, stylization effect, and temporal consistency. These participants come from different backgrounds, making the evaluation less biased towards a certain group of people. Overall, we get 500 votes for images and videos, respectively. Then we calculate the percentage of votes as the \emph{human preference score}. All the evaluations are shown in Tab.~\ref{table1} and Tab.~\ref{table2}.

\begin{figure}[t]
\centering
\includegraphics[width=1.0\textwidth]{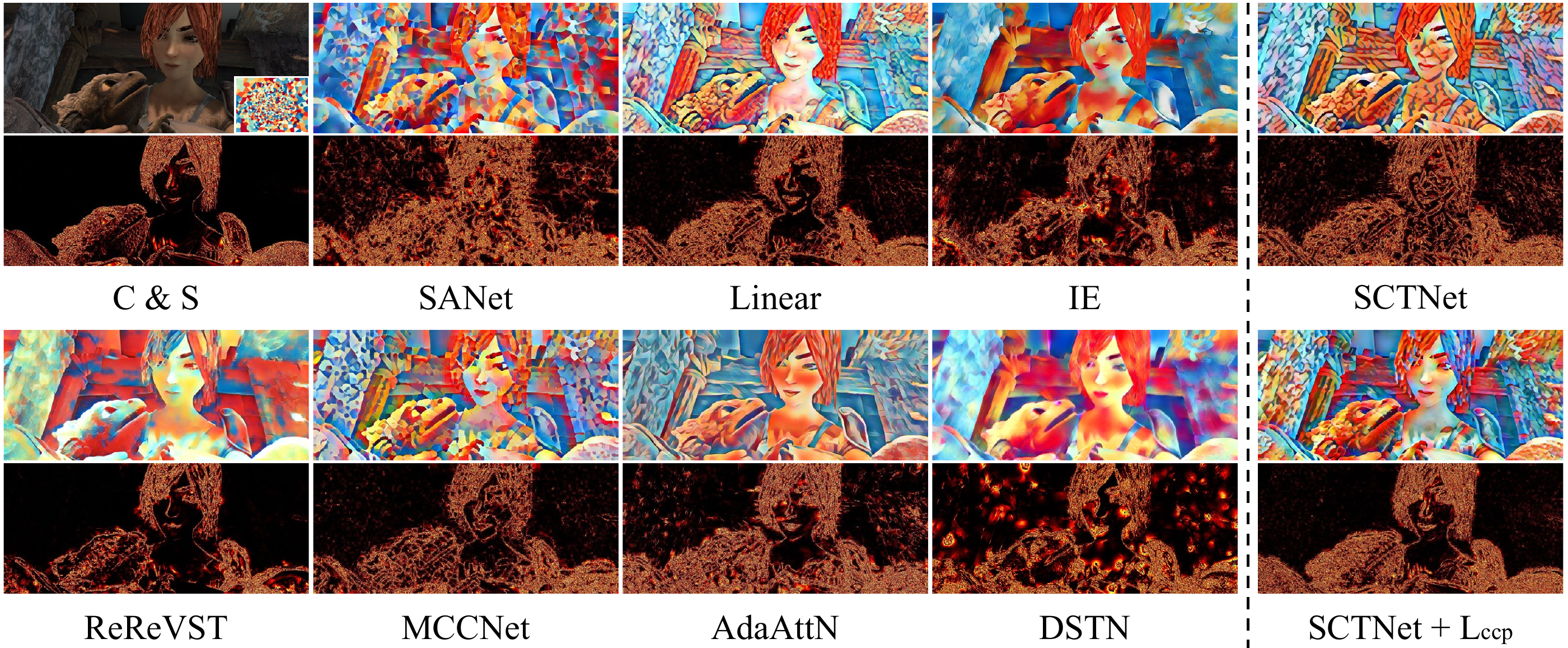} 
\caption{Qualitative comparison of short-term temporal consistency. We compare our method with seven algorithms: SANet~\cite{park2019arbitrary}, Linear~\cite{li2019learning}, IE~\cite{chen2021artistic}, ReReVST~\cite{wang2020consistent}, MCCNet~\cite{deng2020arbitrary}, AdaAttN~\cite{liu2021adaattn}, DSTN~\cite{hong2021domain}. The odd rows show the previous frames. The even rows show the heat-maps of differences between consecutive frames.} 
\label{fig5}
\end{figure}

\subsection{Comparison with Former Methods}
\label{com}
For video and artistic image style transfer, we compare our method with nine algorithms: AdaIN~\cite{huang2017arbitrary}, SANet~\cite{park2019arbitrary}, DSTN~\cite{hong2021domain}, ReReVST~\cite{wang2020consistent}, Linear~\cite{li2019learning}, MCCNet~\cite{deng2020arbitrary}, AdaAttN~\cite{liu2021adaattn}, IE~\cite{chen2021artistic}, $\mathrm{L_{nor}}$~\cite{cheng2021style}, which are the SOTAs of artistic image style transfer. Among these methods, \cite{chen2021artistic,deng2020arbitrary,li2019learning,liu2021adaattn} are also the most advanced single-frame-based video style transfer methods while ReReVST~\cite{wang2020consistent} is the SOTA multi-frames-based method. As for photo-realistic image style transfer, we compare our method with four SOTAs: Linear~\cite{li2019learning}, $\mathrm{WCT^{2}}$~\cite{yoo2019photorealistic}, StyleNAS~\cite{an2019stylenas}, DSTN~\cite{hong2021domain}. Note that among all these mentioned algorithms, Linear~\cite{li2019learning} and DSTN~\cite{hong2021domain} are most relevant to our method, since both of them are capable of transferring artistic and photo-realistic style onto images. We obtain all the test results from the official codes these methods provide.

\begin{figure}[t]
\centering
\includegraphics[width=1.0\textwidth]{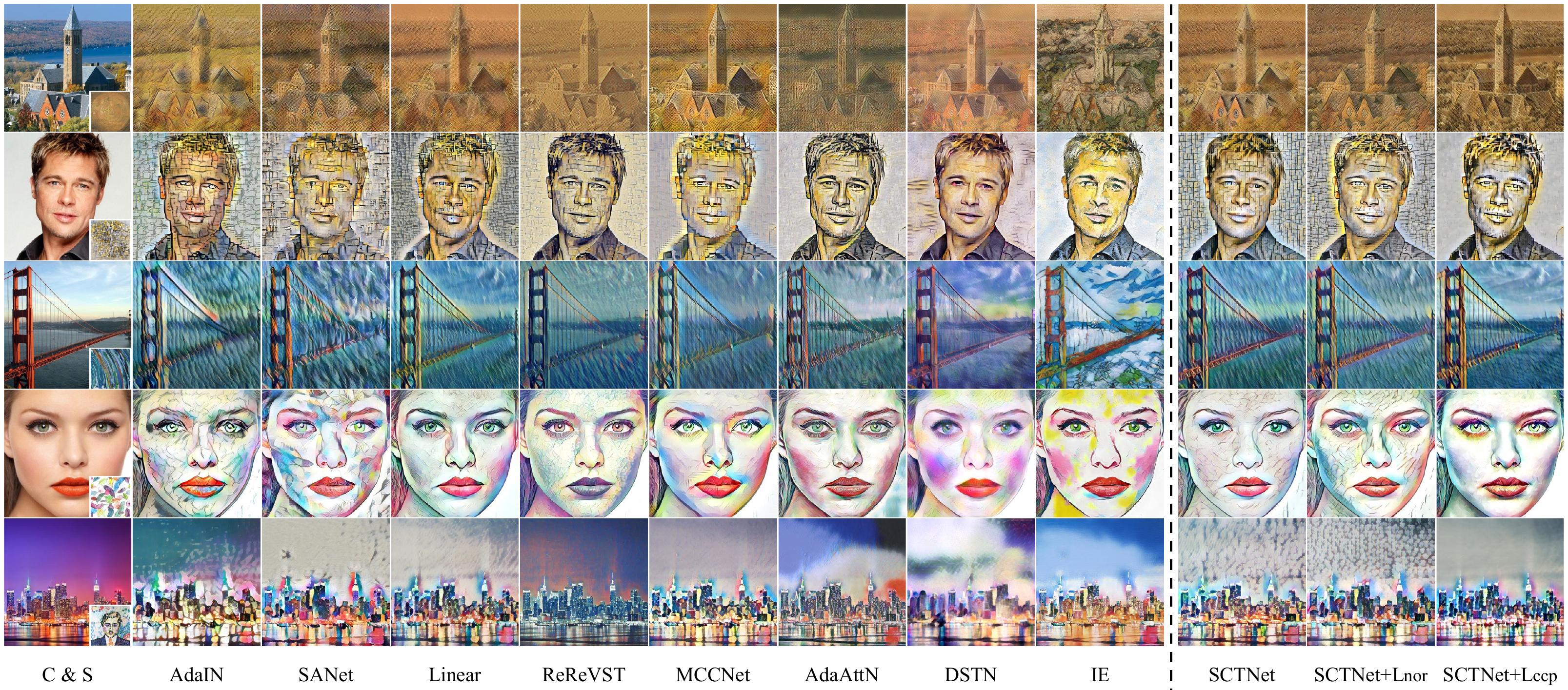} 
\caption{Qualitative comparison of artistic style transfer. We compare our method with nine algorithms: AdaIN~\cite{huang2017arbitrary}, SANet~\cite{park2019arbitrary}, Linear~\cite{li2019learning}, ReReVST~\cite{wang2020consistent}, MCCNet~\cite{deng2020arbitrary}, AdaAttN~\cite{liu2021adaattn}, DSTN~\cite{hong2021domain}, IE~\cite{chen2021artistic}, $\mathrm{L_{nor}}$~\cite{cheng2021style}.}
\label{fig6}
\end{figure}

\vspace{2mm} \noindent\textbf{Video style transfer.} As shown in Tab.~\ref{table1}, our original SCTNet scores the best in SIFID, indicating its superiority in obtaining correct styles. Also, we can see the proposed CCPL improves the temporal consistency a lot with a minor decrease of the SIFID score, when the loss is applied to different methods. And our full model (with CCPL) exceeds all the single-frame methods~\cite{chen2021artistic,deng2020arbitrary,hong2021domain,li2019learning,liu2021adaattn,park2019arbitrary} in both short-term and long-term temporal consistency, which are measured by LPIPS~\cite{zhang2018unreasonable} and temporal loss, and performs on par with the SOTA multi-frame method: ReReVST~\cite{wang2020consistent}. However, our SIFID score exceeds ReReVST~\cite{wang2020consistent} significantly, which is consistent with the results shown in the qualitative comparison (See Fig.~\ref{fig6}). The qualitative comparisons also show the advantage of our CCPL in maintaining short-term (Fig.~\ref{fig5}) temporal consistency of the original video as our heat-map difference is mostly similar to ground-truth. We have another figure in the supplemental file to show the comparison of long-term temporal consistency. In terms of human preference score, our full model also ranks the best, further validating the effectiveness of our CCPL. 

\vspace{2mm} \noindent\textbf{Artistic style transfer.}
As shown in Fig.~\ref{fig6}, AdaIN~\cite{huang2017arbitrary} generates results with severe shape distortion ($e.g.$, house in the $1^{st}$ and bridge in the $3^{rd}$ row) and disarranged texture patterns ($4^{th}$, $5^{th}$ rows). SANet~\cite{park2019arbitrary} also has shape distortion and misses some structural details in its results ($1^{st} \to  3^{rd}$ rows). Linear~\cite{li2019learning} and MCCNet~\cite{deng2020arbitrary} have relatively quite clean outputs. However, Linear~\cite{li2019learning} loses some content details ($1^{st},3^{rd}$ rows), and some results of MCCNet~\cite{deng2020arbitrary} have checkerboard artifacts in local regions (around collar in the $2^{nd}$ row and corner of mouth in the $4^{th}$ row). ReReVST~\cite{wang2020consistent} shows obvious color distortion ($2^{nd}\to 5^{th}$ rows). AdaAttN~\cite{liu2021adaattn} is effective in reducing messy textures but the stylization effect seems to degenerate in some cases ($1^{st}$ row). The results of DSTN~\cite{hong2021domain} have severe obvious distortion ($3^{rd}, 4^{th}$ rows). And the results of IE~\cite{chen2021artistic} are less similar to the original style ($1^{st}, 3^{rd}, 5^{th}$ rows). Our original SCTNet captures accurate style ($2^{nd}, 3^{rd}$ rows), but there are some messy regions in the generated images as well ($4^{th}, 5^{th}$ rows). When adding $\mathrm{L_{nor}}$~\cite{cheng2021style}, some results are even messier ($4^{th}, 5^{th}$ rows). However, with CCPL, the generated results of our full model maintain well the structures of their content sources with vivid and appealing colorization. Besides, this effect is reinforced by its multi-level scheme. Therefore, irregular textures and local color distortions are decreased significantly. It even helps to improve stylization with better preservation of the semantic information of the content sources (as shown in Fig.~\ref{fig9}). 

\begin{figure}[t]
\centering
\includegraphics[width=1.0\textwidth]{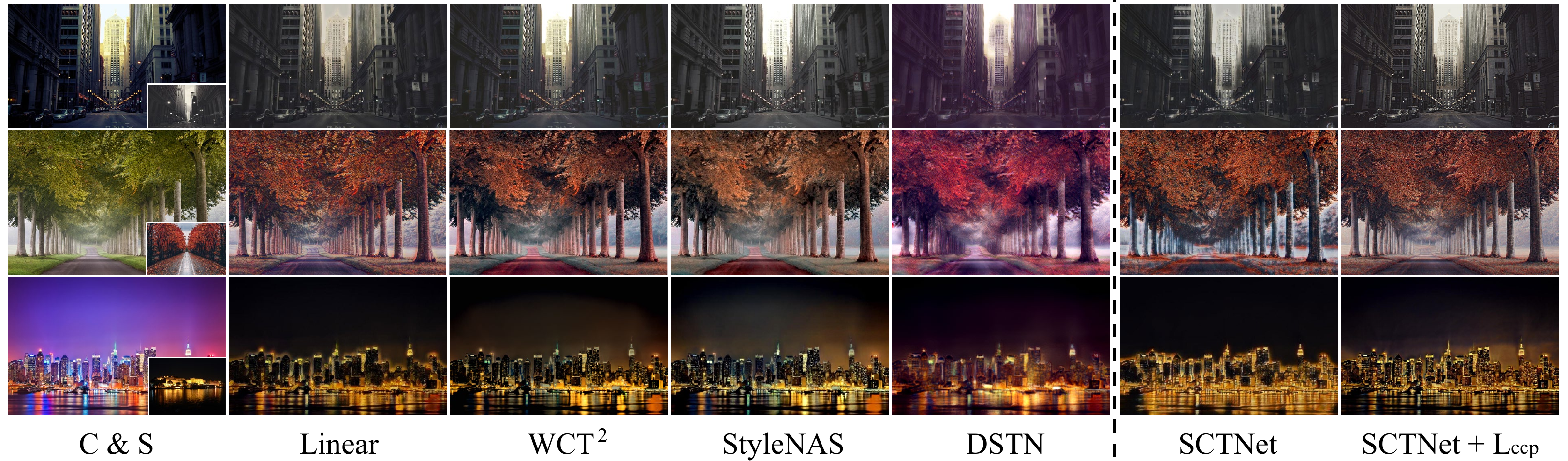} 
\caption{Qualitative comparison of photo-realistic style transfer. We compare our method with four algorithms: Linear~\cite{li2019learning}, $\mathrm{WCT^{2}}$~\cite{yoo2019photorealistic}, StyleNAS~\cite{an2019stylenas} and DSTN~\cite{hong2021domain}.}
\label{fig7}
\end{figure}

\vspace{2mm} \noindent\textbf{Photo-realistic style transfer.} Since CCPL can preserve the semantic information of the content source and significantly reduce local distortions, it is well-suited for the task of photo-realistic style transfer. We make slight changes to SCTNet to enable it for this task: build a shallower encoder by throwing off layers beyond $relu3\_1$, then use feature maps from all three layers to calculate CCPL. As shown in Fig.~\ref{fig7}, Linear~\cite{li2019learning} and DSTN~\cite{hong2021domain} generates results with detail losses (vanished windows in the $3^{rd}$ row). As for $\mathrm{WCT^{2}}$~\cite{yoo2019photorealistic} and StyleNAS~\cite{an2019stylenas}, some results of them show unreasonable color distribution (red road in the $2^{nd}$ row). In comparison, our full model generates results comparable or even better than those SOTAs, with high visual quality and appropriate stylization, which is consistent with the quantitative comparison shown in Tab.~\ref{table2}. 

\vspace{2mm} \noindent\textbf{Efficiency analysis.}
\label{effi}
Our model is quite efficient due to the simple feed-forward architecture of the network and the efficient feature fusion module SCT. We use a single 12GB Titan XP GPU with no other ongoing programs to compare its running speed with other algorithms. Tab.~\ref{table3} shows the average running speed (over 100 independent runs) of different methods on three input image scales. The result suggests that SCTNet surpasses the SOTAs in efficiency at different scales (comparisons for photo-realistic style transfer methods are provided in the supplemental file), indicating the feasibility of our algorithm for real-time use.

\subsection{Ablation Studies}
\label{ablations}
There are several factors relevant to the performance induced by the CCPL: 1) layers to apply the loss; 2) the number of difference vectors sampled each layer; 3) the loss weight ratio with the style loss. Therefore, we conduct several experiments by enumerating the number of CCPL layers from 0 to 4 (start from the deepest layer) and choosing from [16, 32, 64, 128] as the number of sampled combinations to show the impacts of the first two factors. Then we adjust the loss weight ratio between the CCPL and the style loss to manifest which ratio gives the best trade-off between style effects and temporal coherence. To be noted, the stylization score here represents the SIFID score, and the temporal consistency is measured by: ($20 - 10\times \mathrm{LPIPS}-\mathrm{temporal~loss}$) to show the 
escalating trend.

\begin{table}[t]
\centering
\caption{Execution speed comparison (unit: FPS). We use a single 12GB Titan XP GPU for all the execution time testing. OOM denotes the Out-Of-Memory error.}
\begin{tabular}{c|c|c|c|c|c|c|c|c|c}
\toprule[1pt]
    Artistic & Ad~\cite{huang2017arbitrary} & SA~\cite{park2019arbitrary} & LT~\cite{li2019learning} & Re~\cite{wang2020consistent} & MC~\cite{deng2020arbitrary} & AN~\cite{liu2021adaattn} & DN~\cite{hong2021domain} & IE~\cite{chen2021artistic} & Ours \\
    \hline 
    $256\times 256$ & 40.0 & 34.5 & 66.7 & 37.0 & 22.2 & 15.6 & 15.9 & 31.3 & \textbf{77.0}\\
    $512\times 512$ & 12.5 & 14.3 & 18.9 & 13.7 & 8.1 & 12.5 & 4.2 & 13.0 & \textbf{21.7}\\
    $1024\times 1024$ & 2.7 & 2.7 & 4.6 & 2.8 & 1.9 & 2.1 & 1.2 & 2.6 & \textbf{5.0}\\
\bottomrule[1pt]
\end{tabular}
\label{table3}
\end{table}

\begin{figure}[t]
\centering
\includegraphics[width=1.0\textwidth]{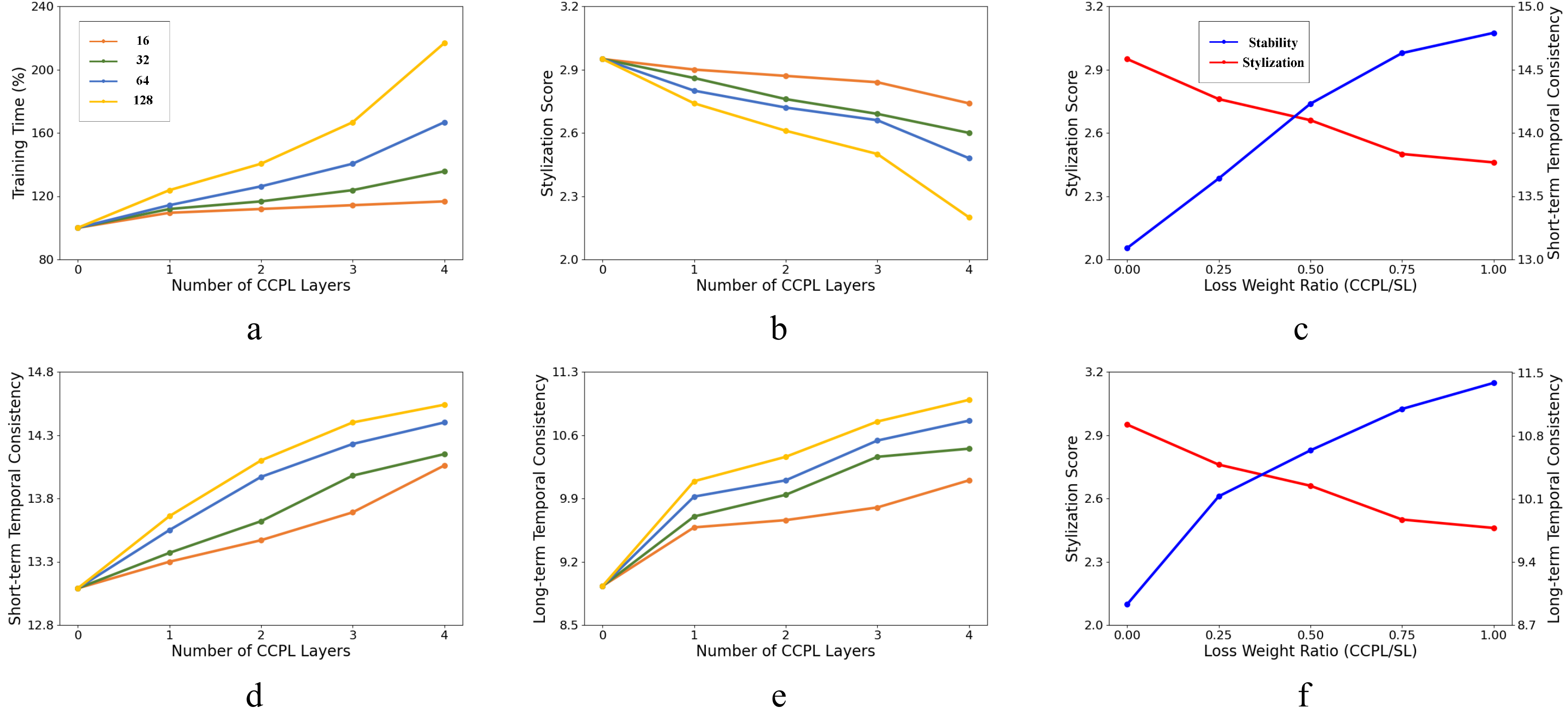} 
\caption{Ablation studies on three factors of the CCPL: 1) layers to apply the loss; 2) the number of vectors sampled each layer; 3) the loss weight ratio with style loss.}
\label{fig8}
\end{figure}

\begin{figure}[ht]
\centering
\includegraphics[width=1.0\textwidth]{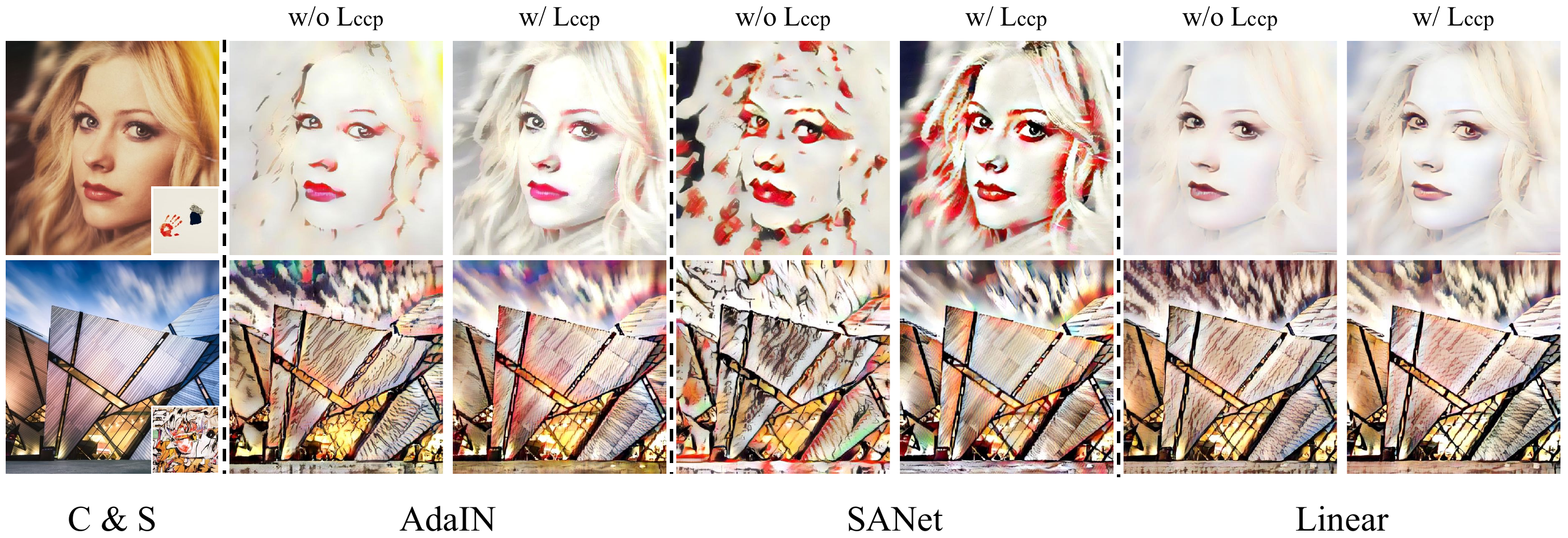} 
\caption{CCPL can be easily applied to other methods, such as AdaIN~\cite{huang2017arbitrary}, SANet~\cite{park2019arbitrary} and Linear~\cite{li2019learning}, to improve visual quality.}
\label{fig9}
\end{figure}

From the sub-figures, we can see that, as the number of CCPL layers increases, the short-term (Fig.~\ref{fig8}$d$) and long-term (Fig.~\ref{fig8}$e$) temporal consistency increases with the reduction of stylization score (Fig.~\ref{fig8}$b$) and greater computation (Fig.~\ref{fig8}$a$). And when the number of CCPL layers increases from 3 to 4, the changes of temporal consistency are minor. In contrast, the computation costs increase significantly, and the stylization effects are much weaker. Therefore, we choose 3 as the default setting for the number of CCPL layers.

As for the number of sampled difference vectors (per layer), the blue lines (64 sampled vectors) in Fig.~\ref{fig8}$d$ \& $e$ are near the yellow lines (128 sampled vectors), which means the performance of these two settings are close on improving temporal consistency. However, sampling 128 difference vectors per layer brings a significantly heavier computation burden and style degeneration. So we sample 64 difference vectors per layer by default. 

The loss weight ratio can also be regarded as a handle to adjust temporal consistency and stylization. Fig.~\ref{fig8}$c$ \& $f$ show the trade-off between temporal consistency and stylization when the loss weight ratio changes. We find 0.5 a good choice for the weight ratio because it gives a good trade-off between temporal consistency improvement and stylization score reduction.
We show the qualitative results of ablation studies on CCPL in the supplemental file and more analysis, such as different sampling strategies in CCPL.

\begin{figure}[t]
\centering
\includegraphics[width=1.0\textwidth]{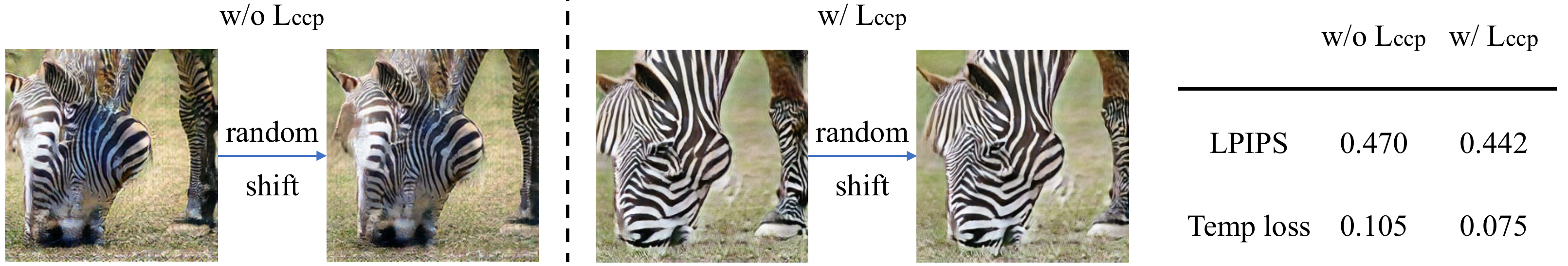} 
\caption{Comparison of applying the CCPL on CUT~\cite{park2020contrastive} with its original model.}
\label{fig10}
\end{figure}

\subsection{Applications}
\label{app}
\noindent\textbf{CCPL on existing methods.}
CCPL is highly flexible and can be plugged into other methods with minor modifications. We apply the proposed CCPL on three typical former methods: AdaIN~\cite{huang2017arbitrary}, SANet~\cite{park2019arbitrary}, Linear~\cite{li2019learning}. All these methods achieve consistent improvements in temporal consistency with only a slight decrease on the SIFID score (see Tab.~\ref{table1} and Fig.~\ref{fig9}). The result reveals the effectiveness and flexibility of the CCPL.

\vspace{2mm} \noindent\textbf{Image-to-image translation.} CCPL can be easily added to other generation tasks like image-to-image translation. We apply our CCPL on a recent image-to-image translation method CUT~\cite{park2020contrastive} and then train the model with the same horse2zebra dataset. The results in Fig.~\ref{fig10} demonstrate that our CCPL improves both the visual quality and temporal consistency. Please refer to the supplemental file for more applications.

\section{Conclusions}
In this work, we propose CCPL to preserve content coherence during style transfer. By contrasting the feature differences of image patches, the loss encourages the difference of patches of the same location in the content and generated images to be similar. Models trained with CCPL achieve a good trade-off between temporal consistency and style effects. We also propose a simple and effective module for aligning second-order statistics of the content feature with style feature. Combining these two techniques, our full model is light and fast while generating satisfying image and video results. Besides, we demonstrate the effectiveness of the proposed loss on other models and tasks, such as image-to-image style transfer, which shows the vast potential of our loss for broader applications.

\noindent\textbf{Acknowledgements} This work was supported by the National Natural Science Foundation of China 62192784.

%
%
\bibliographystyle{splncs04}
\bibliography{egbib}
\end{document}